\documentclass[smallextended]{svjour3}

\smartqed  
\usepackage{graphicx}
\usepackage{booktabs}
\usepackage{xcolor}
\usepackage{siunitx}
\usepackage{url}
\usepackage[misc]{ifsym}
\begin{document}

\title{\texttt{STUDD}: A Student--Teacher Method for Unsupervised Concept Drift Detection}

\titlerunning{Unsupervised Concept Drift Detection}

\author{Vitor~Cerqueira, Heitor~Murilo~Gomes, Albert~Bifet, and Luis~Torgo}

\authorrunning{V. Cerqueira et al.}

\institute{Vitor Cerqueira (\Letter) \at
         Dalhousie University, Halifax, Canada\\
         \email{vitor.cerqueira@dal.ca}
         \and
         Heitor~Murilo~Gomes \at
         University of Waikato, Hamilton, New Zealand\\
         \and
         Albert~Bifet \at
         University of Waikato, Hamilton, New Zealand\\
         Télécom ParisTech, Paris, France
         \and
         Luis Torgo \at
         Dalhousie University, Halifax, Canada\\
         }

\date{Received: date / Accepted: date}

\maketitle

\begin{abstract}

Concept drift detection is a crucial task in data stream evolving environments. Most of state of the art approaches designed to tackle this problem monitor the loss of predictive models. However, this approach falls short in many real-world scenarios, where the true labels are not readily available to compute the loss. 
In this context, there is increasing attention to approaches that perform concept drift detection in an unsupervised manner, i.e., without access to the true labels.
We propose a novel approach to unsupervised concept drift detection based on a student-teacher learning paradigm. Essentially, we create an auxiliary model (student) to mimic the primary model's behaviour (teacher). At run-time, our approach is to use the teacher for predicting new instances and monitoring the \textit{mimicking} loss of the student for concept drift detection. 
In a set of experiments using 19 data streams, we show that the proposed approach can detect concept drift and present a competitive behaviour relative to the state of the art approaches.



\end{abstract}

\section{Introduction}

Learning from time-dependent data is a challenging task due to the uncertainty about the dynamics of real-world environments. 
When predictive models are deployed in environments susceptible to changes, they must detect these changes and adapt themselves accordingly. 
The phenomenon in which the data distribution evolves is referred to as \textit{concept drift}, and a sizeable amount of literature has been devoted to it \cite{gama2014survey}. 

An archetype of concept drift is the interest of users in a service, which typically changes over time \cite{kim2017efficient}. Changes in the environment have a potentially strong negative impact on the performance of models \cite{gama2014survey}. Therefore, it is fundamental that these models can cope with concept drift. That is, to detect changes and adapt to them accordingly.

Concept drift detection and adaptation are typically achieved by coupling predictive models with a change detection mechanism \cite{gomes2019machine}. The detection algorithm launches an alarm when it identifies a change in the data. Typical concept drift strategies are based on sequential analysis \cite{page1954continuous}, statistical process control \cite{gama2004learning}, or monitoring of distributions \cite{bifet2007learning}. When change is detected, the predictive model adapts by updating its knowledge with recent information. A simple example of an adaptation mechanism is to discard the current model and train a new one from scratch. Incremental approaches are also widely used \cite{gomes2017adaptive}.

The input data for the majority of the existing drift detection algorithms is the performance of the predictive model over time, such as the error rate. In many of these detection methods, alarms are signalled if the performance decreases significantly. However, in several real-world scenarios, labels are not readily available to estimate the performance of models. 
Some labels might arrive with a delay or not arrive at all due to labelling costs. 
This is a major challenge for learning algorithms that rely on concept drift detection as the unavailability of the labels precludes their application \cite{gomes2019machine}. 

In this context, there is increasing attention toward unsupervised approaches to concept drift detection. These assume that, after an initial fit of the model, no further labels are available during the deployment of this model in a test set. Most works in the literature handle this problem using statistical hypothesis tests, such as the Kolmogorov-Smirnov test. These tests are applied to the output of the models \cite{vzliobaite2010change}, either the final decision or the predicted probability, or the input attributes \cite{dos2016fast}.

Our goal in this paper is to address concept drift detection in an unsupervised manner. To accomplish this, we propose a novel approach to tackle this problem using a student--teacher learning paradigm called \texttt{STUDD} (\textbf{S}tudent--\textbf{T}eacher approach for \textbf{U}nsupervised \textbf{D}rift \textbf{D}etection). The gist of the idea is as follows. On top of the main predictive model, which we designate as the teacher, we also build a second predictive model, the student. 
Following the literature on model compression \cite{bucilua2006model} and knowledge distillation \cite{hinton2015distilling}, the student model is designed to mimic the behaviour of the teacher. 

Using the student--teacher framework, our approach to unsupervised concept drift detection is carried out by monitoring the student's mimicking loss. The mimicking loss is a function of the discrepancy between the teacher's prediction and student's prediction in the same instance. In summary, we use the student model's loss as a surrogate for the behaviour of the main model. Accordingly, we can apply any state of the art approach in the literature, which considers the loss of a model as the main input, for example, the Page-Hinkley test \cite{page1954continuous}.

When concept drift occurs, it causes changes in the classes' prior probabilities or changes in the class conditional probabilities of the predictor variables. In effect, we hypothesise that these changes disrupt the collective behaviour between the teacher and student models. In turn, this change of behaviour may be captured by monitoring the student model's mimicking loss.

We compared \texttt{STUDD} to several state-of-the-art methods, both unsupervised and supervised ones using 19 benchmark data streams. The results indicate that the proposed method is useful for capturing concept drift. \texttt{STUDD} shows a more conservative behaviour relative to other approaches, which is beneficial in many domains of application.


To summarise, the contributions of this paper are the following:
\begin{itemize}
    \item \texttt{STUDD}: a novel method for unsupervised concept drift detection based on a student--teacher learning approach;
    \item A set of experiments used to validate the proposed method. These include comparisons with state of the art approaches, and an analysis of the different scenarios regarding label availability.
\end{itemize}

The proposed method is publicly available online\footnote{\url{https://github.com/vcerqueira/studd}}. Our implementation is written in Python and is based on the scikit-multiflow framework \cite{montiel2018scikit}. We also remark that this article is an extension of a preliminary work published previously \cite{cerqueira2020unsupervised}.

The rest of this paper is organised as follows. In the next section (Section \ref{sec:pd}), we formally define the problem of concept drift detection in data streams, while in the following section (Section \ref{sec:rw}), we briefly review the literature on the topic of our work. We describe the methodology behind \texttt{STUDD} in Section \ref{sec:methodology}. 
The experiments are reported in Section \ref{sec:experimental_design} we report the results. The results of these are discussed in Section \ref{sec:discussion}. Finally, Section \ref{sec:conclusions} concludes the paper.

\section{Background}\label{sec:pd}

\subsection{Problem Definition}

Let $D(X,y) = \{(X_1, y_1), \dots, (X_t, y_t)\}$ denote a possibly infinite data stream, where each $X$ is a $q$-dimensional array representing the input predictor variables. Each $y$ represents the corresponding output label. We assume that the values of $y$ are categorical. The goal is to use this data set $\{X_i, y_i\}^t_1$ to create a classification model to approximate the function which maps the input $X$ to the output $y$. Let $\mathcal{T}$ denote this classifier. The classifier $\mathcal{T}$ can be used to predict the labels of new observations $X$. We denote the prediction made by the classifier as $\hat{y}_{\mathcal{T}}$.

Many real-world scenarios exhibit a non-stationary nature. Often, the underlying process causing the observations changes in an unpredictable way, which degrades the performance of the classifier $\mathcal{T}$. 
Let $p(X, y)$ denote the joint distribution of the predictor variables $X$ and the target variable $y$. According to Gama et al. \cite{gama2014survey}, concept drift occurs if $p(X, y)$ is different in two distinct points in time across the data stream.
Changes in the joint probability can be caused by changes in $p(X)$, the distribution of the predictor variables or changes in the class conditional probabilities $p(X|y)$ \cite{gao2007general}. These may eventually affect the posterior probabilities of classes $p(y|X)$.

\subsection{Label Availability}\label{sec:label_availability}

When concept drift occurs, the changes need to be captured as soon as possible, so the decision rules of $\mathcal{T}$ can be updated. 
The vast majority of concept drift detection approaches in the literature focus on tracking the predictive performance of the model. If the performance degrades significantly, an alarm is launched and the learning system adapts to these changes. 

\begin{figure}[hbt]
\centering
\includegraphics[width=.75\textwidth]{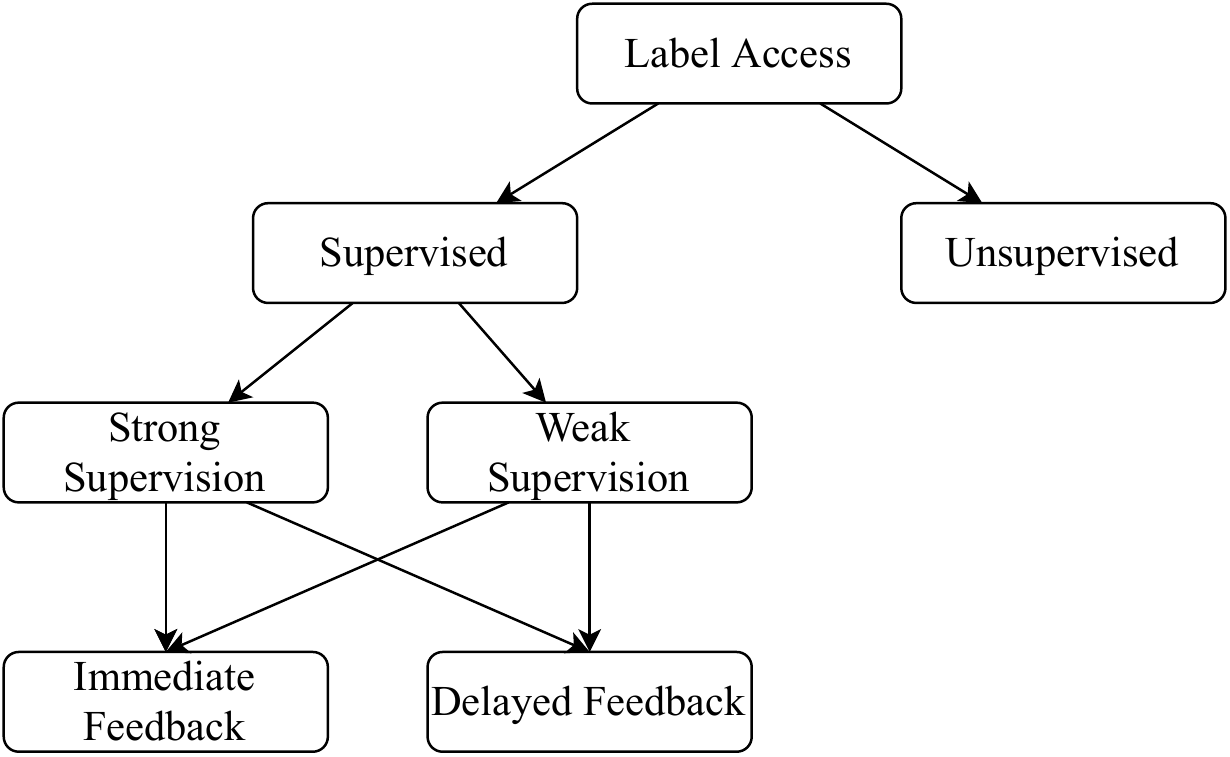}
\caption{The distinct potential scenarios regarding label access after the initial fit of the model (adapted from Gomes et al. \cite{gomes2017adaptive}).}
\label{fig:labels}
\end{figure}

The problem with these approaches is that they assume that the true labels are readily available after prediction. In reality, this is rarely the case. In many real-world scenarios, labels can take too long to be available, if ever. If labels do eventually become available, often we only have access to a part of them. This is due to, for example, labelling costs. The different potential scenarios when running a predictive model are depicted in Figure \ref{fig:labels}.

Precisely, a predictive model is built using an initial batch of training data, whose labels are available. When this model is deployed in a test set, concept drift detection is carried out in an unsupervised or supervised manner. 

In unsupervised scenarios, no further labels are available to the predictive model. Concept drift detection must be carried out using a different strategy other than monitoring the loss. For example, one can track the output probability of the models \cite{vzliobaite2010change} or the unconditional probability distribution $p(X)$ \cite{kuncheva2004classifier}. 

Concept drift detectors have access to labels when the scenario is supervised. On the one hand, the setting may be either strongly supervised or weakly supervised \cite{zhou2018brief}. In the former, all labels become available. In the latter, the learning system only has access to a part of the labels. This is common in applications which data labelling is costly. On the other hand, labels can arrive immediately after prediction, or they can arrive with some delay. In some domains, this delay may be too large, and unsupervised approaches need to be adopted.

In this paper, we address concept drift detection from an unsupervised perspective. In this setting, we are restricted to use $p(X)$ to detect changes, as the probability of the predictor variables is not conditioned on $y$.

\section{Related Research}\label{sec:rw}

In this section, we briefly review previous research related to our work. We split this review into two parts. In the first part, we overview approaches for concept drift detection, giving particular emphasis to unsupervised approaches.
The second part addresses model compression and the related work on the student--teacher learning approach, which is the basis of the proposed method.

\subsection{Concept Drift Detection}
Concept drift can occur in mainly three different manners: suddenly, in which the current concept is abruptly replaced by a new one; gradually, when the current concept slowly fades; and reoccurring, in which different concepts are prevalent in distinct time intervals (for example, due to seasonality). A variation of gradual concept drifts are incremental drifts, which are extremely difficult to detect as they consist of many concepts that continually evolve. 

We split concept drift detection into two dimensions: supervised and unsupervised. The supervised type of approaches assumes that the true labels of observations are available after prediction. Hence, they use the error of the model as the main input to their detection mechanism. On the other hand, unsupervised approaches preclude the use of the labels in their techniques.

\subsubsection{Supervised Approaches}

Plenty of error-based approaches have been developed for concept drift detection. These usually follow one of three sort of strategies: sequential analysis, such as the Page-Hinkley test (PHT) \cite{page1954continuous}; statistical process control, for example the Drift Detection Method (DDM) \cite{gama2004learning} or the Early Drift Detection Method (EDDM) \cite{baena2006early}; and distribution monitoring, for example the Adaptive Windowing (ADWIN) approach \cite{bifet2007learning}.

\subsubsection{Unsupervised Approaches}\label{sec:rw_unsupervised}

\begin{figure}[hbt]
\centering
\includegraphics[width=.9\textwidth]{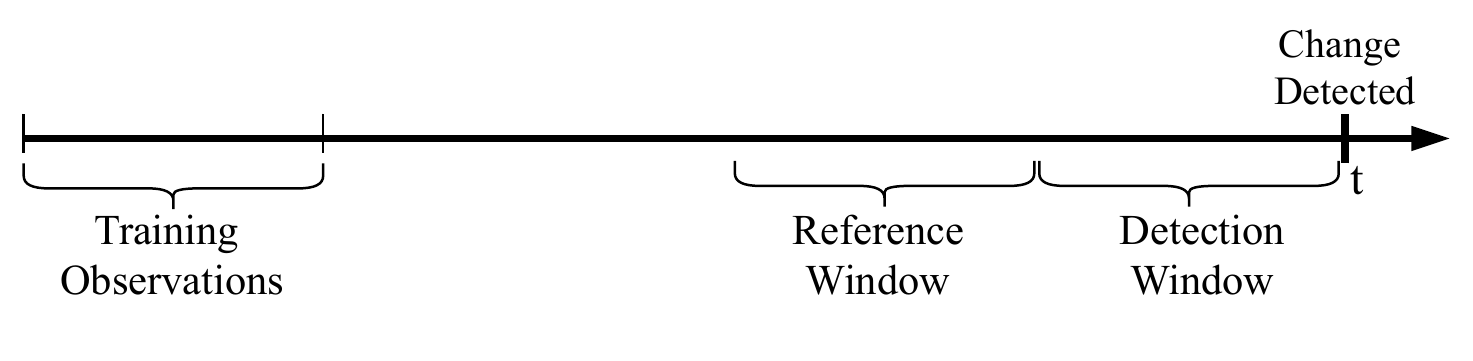}
\caption{Detecting changes using a sliding reference window. Change occurs at time t if the reference window is statistically different than the detection window.}
\label{fig:output_tracker_sliding}
\end{figure}

\begin{figure}[hbt]
\centering
\includegraphics[width=.9\textwidth]{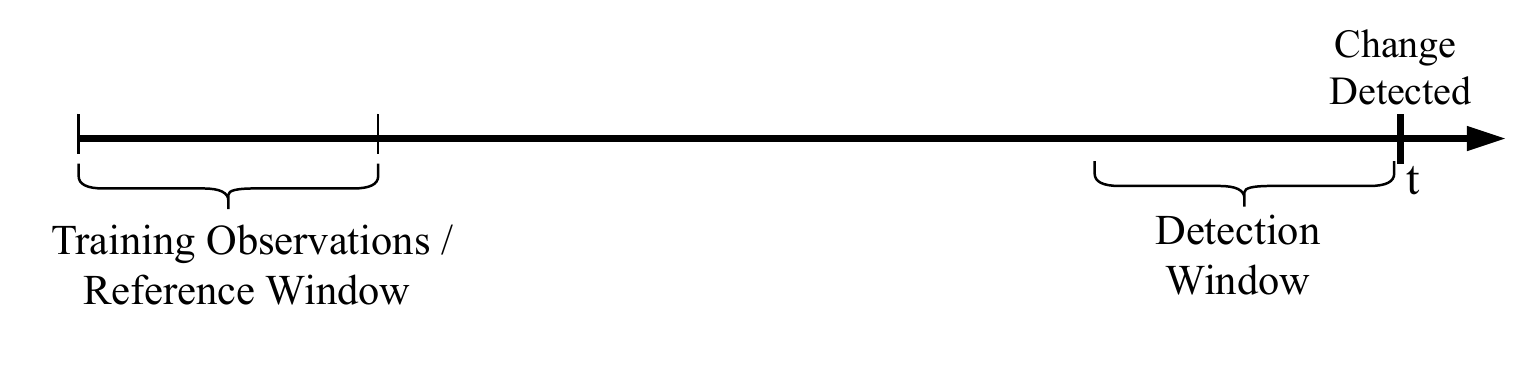}
\caption{Detecting changes using a fixed reference window.}
\label{fig:output_tracker_fixed}
\end{figure}

Although the literature is scarce, there is an increasing interest in approaches which try to detect drift without access to the true labels. \v{Z}liobaite \cite{vzliobaite2010change} presents a work of this type. She proposed the application of statistical hypothesis testing to the output of the classifier (either the probabilities or the final categorical decision). The idea is to monitor two samples of one of these signals. One sample serves as the reference window, while the other represents the detection window. When there is a statistical difference between these, an alarm is launched. 
This process can be carried out using a sliding reference window (c.f. Figure \ref{fig:output_tracker_sliding}) or a fixed reference window (c.f. Figure \ref{fig:output_tracker_fixed}).

In a set of experiments, \v{Z}liobaite shows that concept drift is detectable using this framework. The hypothesis tests used in the experiments are the two-sample Kolmogorov-Smirnov test, the Wilcoxon rank-sum test, and the two-sample t-test.

Reis et al. \cite{dos2016fast} follow a strategy similar to \v{Z}liobaite \cite{vzliobaite2010change}. They propose an incremental version of the Kolmogorov-Smirnov test and use this method to detect changes without using any true labels. 
However, they focus on tracking the attributes rather than the output of the predictive model. Specifically, they use a fixed window approach (c.f. Figure \ref{fig:output_tracker_fixed}) to monitor the distribution of each attribute. If a change is detected in any of these attributes, a signal for concept drift is issued.

In the same line of research, Yu et al. \cite{yu2018request} apply two layers of hypothesis testing hierarchically. Kim et al. \cite{kim2017efficient} also apply a windowing approach. Rather than monitoring the output probability of the classifier, they use a confidence measure as the input to drift detectors.

Pinto et al. \cite{pinto2019automatic} present an automatic framework for monitoring the performance of predictive models. Similarly to the above-mentioned works, they perform concept drift detection based on a windowing approach. The signal used to detect drift is computed according to a mutual information metric, namely the Jensen-Shannon Divergence \cite{lin1991divergence}. The window sizes and threshold above which an alarm is launched is analysed, and the approach is validated in real-world data sets. The interesting part of the approach by Pinto et al. \cite{pinto2019automatic} is that their method explains the alarms. This explanation is based on an auxiliary binary classification model. The goal of applying this model is to rank the events that occurred in the detection window according to how these relate to the alarm. These explanations may be crucial in sensitive applications which require transparent models.

G{\"o}z{\"u}a{\c{c}}{\i}k et al. \cite{gozuaccik2019unsupervised} also develop an auxiliary predictive model for unsupervised concept drift detection, which is called D3 (for (Discriminative Drift Detector). The difference to the work by Pinto et al. \cite{pinto2019automatic} is that they use this model for detecting concept drift rather than explaining the alarms.

\subsection{Student--Teacher Learning Approach}

Model compression, also referred to as student-teacher learning, is a technique proposed by Bucilu\v{a} et al. \cite{bucilua2006model}. The goal is to train a model, designated as a student, to mimic the behaviour of a second model (the teacher). 
To perform model compression, the idea is to first retrieve the predictions of the teacher in observations not used for training (e.g. a validation data set). Then, the student model is trained using this set of observations, where the explanatory variables are the original ones, but the original target variable is replaced with the predictions of the teacher. 
The authors use this approach to compress a large ensemble (the teacher) into a compact predictive model (the student).

Bucilu\v{a} et al. \cite{bucilua2006model} use the ensemble selection algorithm \cite{caruana2004ensemble} as the teacher and a neural network as the student model and address eight binary classification problems. Their results show that the compressed neural network performs comparably with the teacher while being ``1000 times smaller and 1000 times faster''. Moreover, the compressed neural network considerably outperforms the best individual model in the ensemble used as the teacher.

Hinton et al. \cite{hinton2015distilling} developed the idea of model compression further, denoting their compression technique as knowledge distillation. Distillation works by softening the probability distribution over classes in the softmax output layer of a neural network.
The authors address an automatic speech recognition problem by distilling an ensemble of deep neural networks into a single and smaller deep neural network. 

Both Bucilu\v{a} et al. \cite{bucilua2006model} and Hinton et al. \cite{hinton2015distilling},  show that combining the predictions of the ensemble leads to a comparable performance relative to a single compressed model. 

While our concerns are not about decreasing the computational costs of a model, we can leverage model compression approaches to tackle the problem of concept drift detection. Particularly, by creating a student model which mimics the behaviour of a classifier, we can perform concept drift detection using the loss of the student model. Since this loss is not conditioned on the target variable $y$, concept drift detection is carried out in an unsupervised manner.

This paper significantly extends a previously published paper \cite{cerqueira2020unsupervised}. The experiments are completely different. While in the previous work we validated \texttt{STUDD} using synthetic drifts based on two data sets, we now focus on a realistic evaluation scenario based on 19 benchmark data streams. We also include other state of the art approaches in the experiments.

\section{Methodology}\label{sec:methodology}

In this section we describe \texttt{STUDD}, the proposed approach to unsupervised concept drift detection.
\texttt{STUDD} is split into two steps: an initial offline stage, which occurs during the training of the data stream classifier (Section \ref{sec:stage1}); and an online stage, when the method is applied for change detection (Section \ref{sec:stage2}).

\subsection{Stage 1: Student--Teacher Training}\label{sec:stage1}

\begin{figure}[hbt]
\centering
\includegraphics[width=\textwidth]{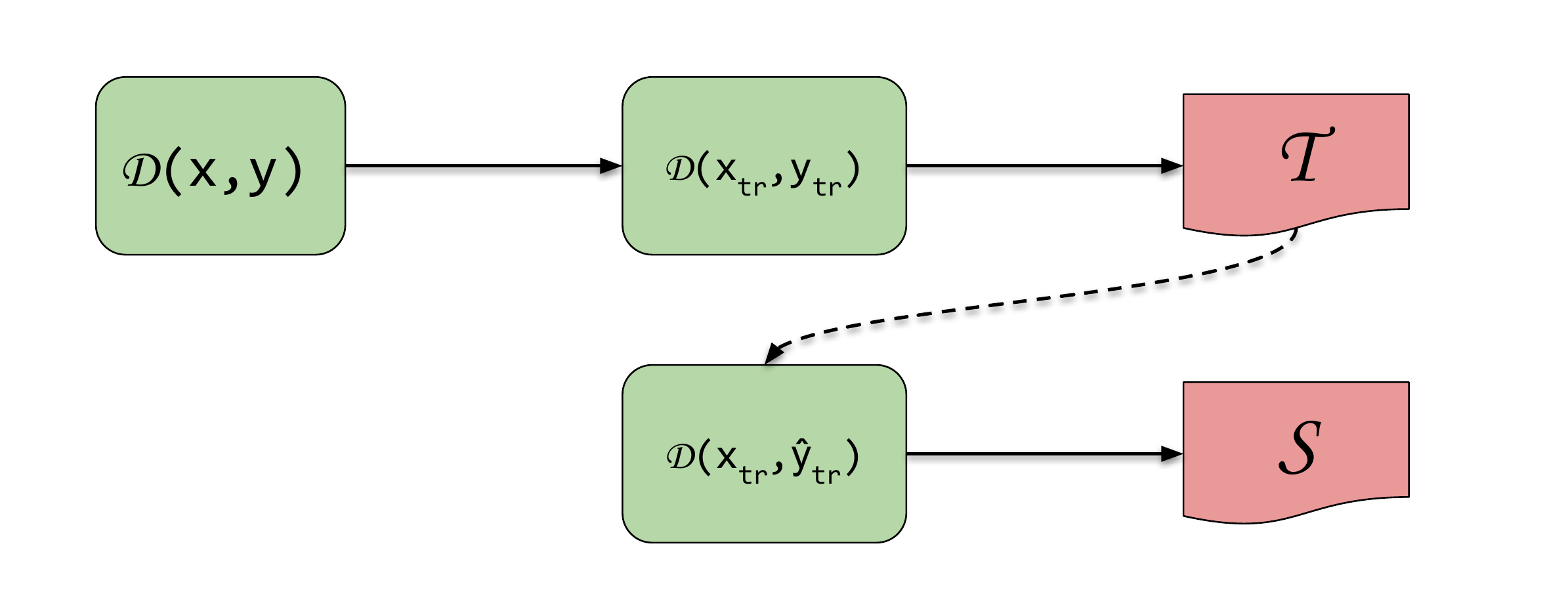}
\caption{Fitting the teacher ($\mathcal{T}$) and student ($\mathcal{S}$) models using an initial batch of training observations.}
\label{fig:scheme_fitting}
\end{figure}

The first stage of the proposed approach refers to the training of the predictive models. This process is illustrated in Figure \ref{fig:scheme_fitting}. A batch of training observations is retrieved from the source data stream $\mathcal{D}$. These observations ($\mathcal{D}(X_{tr}, y_{tr})$) are used to train the classifier $\mathcal{T}$. This is the predictive model to be deployed in the data stream.

After creating $\mathcal{T}$, we carry out a student--teacher approach in which $\mathcal{T}$ acts as the teacher. First, $\mathcal{T}$ is used to make predictions on the training set. This leads to a new training data set, in which the targets $y_{tr}$ are replaced with the predictions of $\mathcal{T}$, $\hat{y}_{\{\mathcal{T},tr\}}$. Finally, the student model $\mathcal{S}$ is trained using the new data set. Essentially, the student model $\mathcal{S}$ is designed to mimic the behavior of the teacher $\mathcal{T}$.

It might be argued that using the same instances to train both the teacher and the student models leads to over-fitting. However, Hinton et al. \cite{hinton2015distilling} show that this is not a concern. 

The student-teacher learning paradigm is at the core of model compression \cite{bucilua2006model} or knowledge distillation \cite{hinton2015distilling} methods. These approaches aim at compressing a model with a large number of parameters (teacher), such as an ensemble or a deep neural network, into a more compact model (student) with a comparable predictive performance. Accordingly, the student model is deployed in the test set, while the teacher is not used in practice due to high computational costs.

Conversely, our objective for using a student-teacher strategy is different. We regard the student model $\mathcal{S}$ as a model which is able to predict the behavior of the teacher model $\mathcal{T}$, i.e., what the output of $\mathcal{T}$ will be for a given input observation. Moreover, it is important to remark that, in our methodology, both student and teacher models are applied in the test phase.

\subsection{Stage 2: Change Detection}\label{sec:stage2}

The second stage of the proposed method refers to the change detection process. As we have described before, the state-of-the-art concept drift detection methods take the loss of predictive models as their primary input. Since we assume that labels are unavailable, we cannot compute the model's loss $\mathcal{T}$. This precludes the typical application of state-of-the-art change detection approaches to unsupervised concept drift detection. 

\begin{figure}[hbt]
\centering
\includegraphics[width=\textwidth]{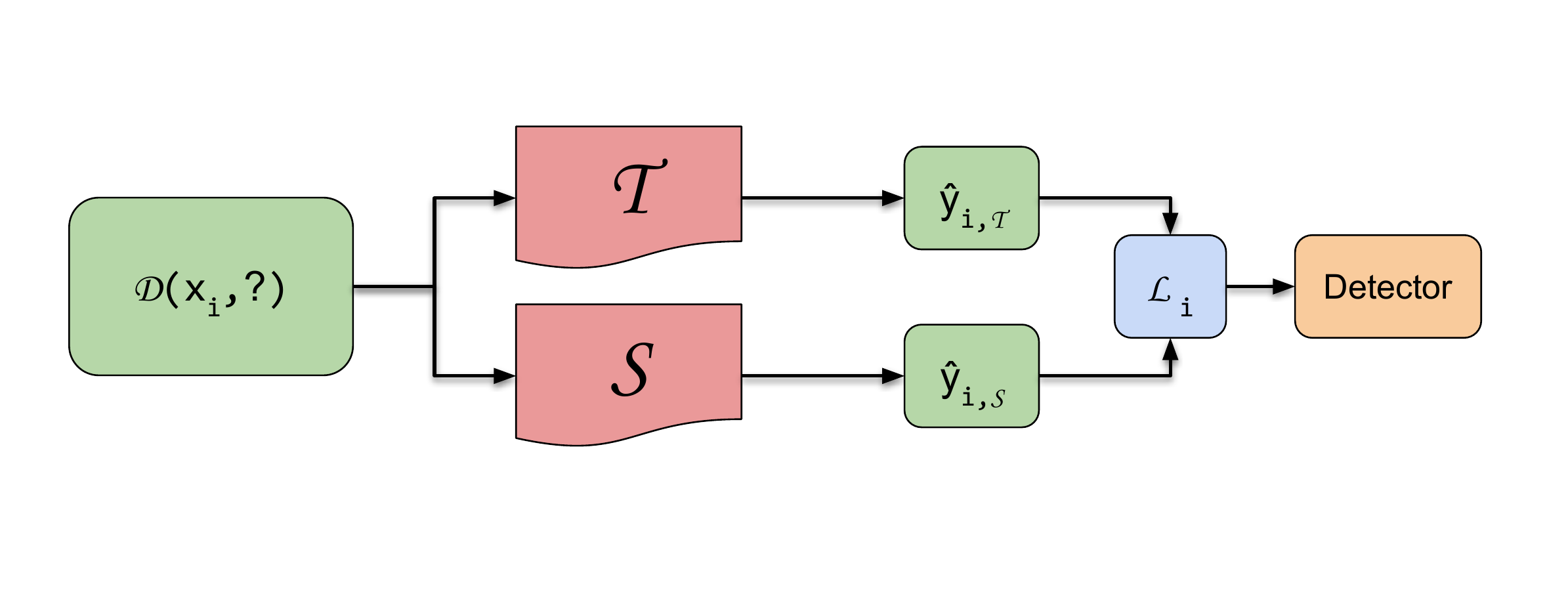}
\caption{The concept drift detection process of \texttt{STUDD}. We retrieve the predictions from both models for a new observation. A function of the discrepancy of these predictions, which is independent from the true labels, is given as input to a state of the art detection model.}
\label{fig:scheme_detection}
\end{figure}

However, we can compute the student model's loss, which is independent of the true labels.
The loss of the student is quantified according to the discrepancy between the prediction of $\mathcal{T}$ ($\hat{y}_{\mathcal{T}}$) and the prediction of $\mathcal{S}$ about $\hat{y}_{\mathcal{T}}$ ($\hat{y}_{\mathcal{S}}$). Accordingly, the loss of $\mathcal{S}$ is defined as $L(\hat{y}_{\mathcal{T}}, \hat{y}_{\mathcal{S}})$, where $L$ is the loss function (e.g. the error rate). 

Therefore, our approach to concept drift detection uses a state of the art detector, such as the Page-Hinkley test \cite{page1954continuous}. However, the main input to this detector is the student model's error, rather than the teacher model's error. This process is depicted in Figure \ref{fig:scheme_detection}. For a given input observation $x_i$, we obtain the prediction from the models $\mathcal{T}$ and $\mathcal{S}$. Then, a function of the discrepancy between these predictions is given as input to the detection model. 

When concept drift occurs, it potentially causes changes in the posterior probability of classes, $p(y|X)$. Thus, we hypothesise that such changes will also potentially affect the joint behaviour between student and teacher models. This effect will then be reflected on the student's imitation error, and the underlying change detection mechanism can capture it.

In effect, the teacher model is deployed in the data stream and used to make predictions on the upcoming observations. For concept drift detection, we track the error of the student model.

\section{Empirical Experiments}\label{sec:experimental_design}

This section details the experiments carried out to validate the proposed approach to unsupervised concept drift detection.
We start by describing the research questions we aim at answering (\ref{sec:research_questions}), followed by a brief description of the data streams used in the experiments (\ref{sec:data}). 
Afterwards, we explain the workflow used to analyse each approach under comparison (\ref{sec:workflow_experiments}), and in the respective evaluation scheme (\ref{sec:evaluation}). Finally, we describe the methods used to compare \texttt{STUDD} with (\ref{sec:methods}), and detail the value of important parameters (\ref{sec:parameter_setup}).

\subsection{Research Questions}\label{sec:research_questions}

We designed a set of experiments to answer the following research questions:
\begin{itemize}
    \item \textbf{RQ1}: Is \texttt{STUDD} able to detect concept drift?
    
    \item \textbf{RQ2}: What is the performance of \texttt{STUDD} for concept drift detection relative to state of the art approaches? These include both unsupervised and supervised ones;
    
    \item \textbf{RQ3}: When, in terms of label availability scenarios, is \texttt{STUDD} beneficial relative to a supervised approach?
\end{itemize}

\subsection{Data Sets}\label{sec:data}

\begin{table}[!thb]
	\centering
	\caption{Data streams used in the experiments. The shape column describes the dimensionality of the data in the form (number of rows $\times$ number of columns).}		
	\resizebox{\textwidth}{!}{%
	\begin{tabular}{lp{5cm}rr}
	\toprule
	\textbf{Data Stream} & \textbf{Description} & \textbf{Shape} & \textbf{\# Classes}\\ 
	\midrule
	    Electricity & \strut Price direction of electricity market in Australia & 45.312 $\times$ 9 & 2 \\
	    
	    CoverType & \strut Forest cover type from the US Forest Service & 581.012 $\times$ 55 & 5\\
	    
	    Poker & \strut Poker hands drawn from a deck of 52 cards & 1.025.010 $\times$ 12 & 10\\
	    
	    Gas & \strut Gas measurements from chemical sensors & 13.910 $\times$ 17 & 6\\
	    
	    Luxembourg & \strut Survey concerning the internet usage (high or low) from 2002 to 2007 & 1.901 $\times$ 31 & 2\\
	    
	    Ozone & Air measurements concerning ozone levels & 2.574 $\times$ 32 & 2\\
	    
	    Sensors & Sensor identification from environmental data &  2.219.803 $\times$ 5 & 54\\
	    
	    Powersupply & Hour identification from power supply data from an Italian electricity company &  29.928 $\times$ 2 & 24\\
	    
	    Rialto & Building identification near from processed images taken in the Rialto bridge in Venice & 82.250 $\times$ 27 & 10 \\
	    
	    Outdoor & Object identification from images taken outdoor under varying lighting conditions (sunny and cloudy) & 4.000 $\times$ 21 & 40 \\
	    
	    Keystroke & User identification from typing rhythm of an expression & 1.600 $\times$ 10 & 4\\
	    
	    NOAA & Rain detection from weather measurements collected over 50 years & 18.159 $\times$ 8 & 2\\
	    
	    Bike & Count of rental bikes (high or low) from a bike-sharing system & 17.378 $\times$ 5 & 2\\
	    
	    Arabic & Digit identification from audio (in arabic) features & 8.800 $\times$ 28 & 10\\
	    
	    ArabicShuffled & Similar to \textit{Arabic} but observations are shuffled by gender to enhance concept drift (c.f. \cite{dos2016fast}) & 8.800 $\times$ 28 & 10\\
	    
	    Insects & Identification of the specimen of a flying insect that is passing through a laser & 5.325 $\times$ 50 & 5\\
	    
	    InsectsAbrupt & Similar to \textit{Insects}, but abrupt drift is introduced in the feature space & 5.325 $\times$ 50 & 5\\
	    
	    Posture & Movement identification from sensors carried by different people & 164.859 $\times$ 4 & 11 \\
	    
	    GMSC & Credit scoring data set (\textit{Give me some credit}) & 150.000 $\times$ 11 & 2 \\
	    
		\bottomrule    
	\end{tabular}%
	}
	\label{tab:data}
\end{table}

We used 19 benchmark data streams to answer the above research questions and validate the applicability of \texttt{STUDD}.
These data sets include the following data streams: Electricity \cite{harries1999splice}, forest cover type \cite{blackard1999comparative}, Poker \cite{cattral2002evolutionary}, Gas \cite{vergara2012chemical}, Luxembourg \cite{vzliobaite2011combining}, Ozone \cite{dua2017uci}, Power supply \cite{zhu2010stream}, Rialto \cite{losing2016knn}, Outdoor \cite{losing2015interactive}, Keystroke \cite{souza2015data}, NOAA \cite{ditzler2012incremental}, Bike \cite{fanaee2014event}, Arabic \cite{hammami2010improved}, Arabic with shuffled observations as per Reis et al. \cite{dos2016fast}, Insects \cite{de2013classification}, Insects with artificial abrupt concept drift \cite{dos2016fast}, Posture \cite{kaluvza2010agent}, and GMSC \cite{gomes2017adaptive}. These are briefly described in Table \ref{tab:data}. In order to speed up computations, we truncated the sample size of all data streams to 150.000 observations. These data sets are commonly used as benchmarks for mining data streams. We retrieved them from an online repository for data streams \cite{souza2020challenges}, or the repository associated with two previous works related to data streams \footnote{\url{https://github.com/denismr/incremental-ks}}\textsuperscript{,}\footnote{\url{https://github.com/hmgomes/AdaptiveRandomForest}}.

\subsection{Workflow of Experiments}\label{sec:workflow_experiments}

We designed the experiments according to a batch setup, split into an offline stage and an online stage.

In the offline stage, we train the main classifier $\mathcal{T}$ to be deployed in the data stream using an initial batch of W observations. We also carry out any task-specific to the underlying drift detection approach. For example, in the case of the proposed approach, we also train the student model $\mathcal{S}$. 

The online stage starts when the classifier $\mathcal{T}$ is deployed in the data stream. For each new observation $x_i$, the classifier $\mathcal{T}$ makes a prediction $\hat{y}_i$. Meanwhile, the underlying detection mechanism uses the available data (e.g. $x_i$, $\hat{y}_i$) to monitor the classifier's behaviour. If the detection mechanism detects a change, it launches an alarm and the classifier $\mathcal{T}$ is adapted with recent information.

\begin{figure}[hbt]
\centering
\includegraphics[width=.9\textwidth]{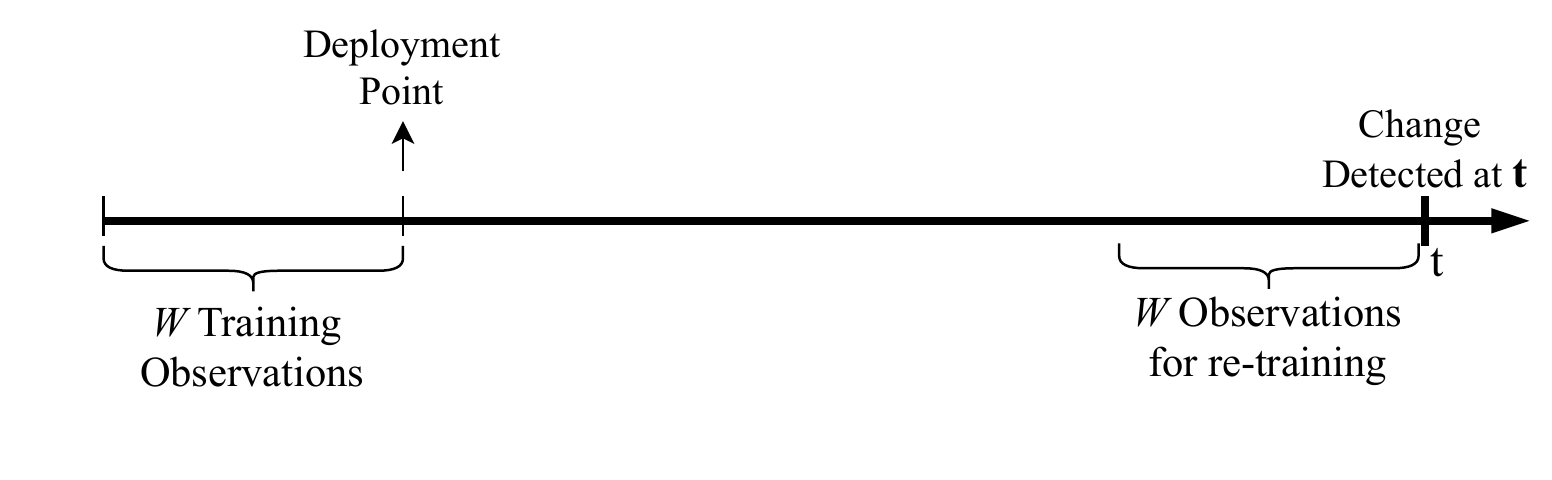}
\caption{The workflow for applying and evaluating each method under comparison.}
\label{fig:workflow_experiments}
\end{figure}

The adaptation mechanism adopted in this work is based on a re-training procedure. The current model is discarded, and a new model is re-trained using the latest W observations. This workflow is depicted in Figure \ref{fig:workflow_experiments}. We remark that, in the case of \texttt{STUDD}, the student model is also updated as described.

\subsection{Evaluation}\label{sec:evaluation}

Our goal is to evaluate the performance of the concept drift detection mechanisms. We focus on unsupervised scenarios, in which the true labels are not readily available. 
However, for evaluation purposes, we use the labels to assess the quality of change detectors.
We aim at measuring a trade-off between predictive performance and the number of alarms issued by the detector. The alarms represent a cost as they trigger the retrieval (and annotation) of a batch of observations.

We evaluate each approach from two dimensions:
\begin{itemize}
    \item Predictive performance: We measure the quality of the classifier according to the Cohen's Kappa statistic \cite{cohen1960coefficient}, which is a common metric used to evaluate classification models;
    
    \item Annotation costs: We assume that we are working in environments where labels are scarce and costly to obtain. Therefore, as explained before, each concept drift signal triggers a request for an additional batch of labels. This request is expensive to the user, and an important metric to minimize. We account for this problem by measuring the ratio of labels with respect to the complete length of the data stream used by the respective approach.
\end{itemize}

\noindent Ideally, the optimal approach maximizes predictive performance and minimizes the amount of labels requested.

\subsection{Methods}\label{sec:methods}

Besides the proposed method we include nine other approaches in our experimental setup. These can be described as follows.

We include the following two baselines:
\begin{itemize}
    \item \texttt{BL-st}: A static baseline, which never adapts to concept drift. In practice, a model (Random Forest) is fit using the training observations and used to predict the subsequent ones remaining in the data stream. Accordingly, the model may become outdate due to concept drift, but incurs in minimal labelling costs;
    
    \item \texttt{BL-ret}: A baseline which follows the opposite strategy to \texttt{BL-st}; it retrains the predictive model after every $W$ observations. The model is always up to date, but at the price of high labelling costs.
\end{itemize}

In terms of unsupervised methods, which do not use any true labels, we apply the following state of the art approaches:

\begin{itemize}
    \item \texttt{Output Sliding}  (\texttt{OS}): A method that tracks the output of the predictive model using a \textbf{sliding} reference window as described by {\v{Z}}liobaite \cite{vzliobaite2010change}. Figure \ref{fig:output_tracker_sliding} shows the workflow of this approach. According to previous studies \cite{vzliobaite2010change,dos2016fast}, we apply the Kolmogorov-Smirnov test to assess whether or not change occurs (c.f. Section \ref{sec:rw_unsupervised} for more details); 
    
    \item \texttt{Output Fixed} \texttt{OF}: A similar approach to \texttt{OS}, but which tracks the output of the predictive model using a \textbf{fixed} reference window. This approach is depicted in Figure \ref{fig:output_tracker_fixed}. We also apply the Kolmogorov-Smirnov test in this case.

    \item \texttt{Feature Fixed} (\texttt{Feature Fixed}): The method described by Reis et al. \cite{dos2016fast}, which instead of tracking the output of predictive models (such as \texttt{OS} or \texttt{OF}), it tracks the values of features. Following Reis et al. \cite{dos2016fast}, if a change is detected in any of the features using the Kolmogorov-Smirnov test, the predictive model is adapted.
    
\end{itemize}

We also include the following supervised approaches in our experiments.
These assume some level of access to the true labels. While they may not be applicable in some scenarios where labels are difficult to acquire, they are important benchmarks for comparisons.

\begin{itemize}
    \item \texttt{Strongly Supervised} (\texttt{SS}): We apply the standard concept drift detection procedure which assumes that all the true labels are immediately available after making a prediction. This can be regarded as the gold standard. The term \textit{strong} means refers to the fact that all labels are available during testing \cite{zhou2018brief};
    
    \item \texttt{Weakly Supervised} (\texttt{WS}): In many real-world scenarios, particularly in high-frequency data streams, data labelling is costly. Hence, predictive models can only be updated using a part of the entire data set. This process is commonly referred to as weakly supervised learning \cite{zhou2018brief}.
    We simulate a weakly supervised scenario in our experiments. Accordingly, predictive models only have access to \textit{l\_access}\% of the labels. In other words, after a model predicts the label of a given instance, the respective label is immediately available with a \textit{l\_access}\% probability;
    
    \item \texttt{Delayed Strongly Supervised} (\texttt{DSS}): Labels can take some time to be available. We study this aspect by artificially delaying the arrival of the labels by \textit{l\_delay} instances. After a label becomes available, the respective observation is used to update the change detection model;
    
    \item \texttt{Delayed Weakly Supervised} (\texttt{DWS}): We combine the two previous scenarios. In the \texttt{DWS} setup, only \textit{l\_access}\% of the labels are available. Those which are available arrive with a delay of \textit{l\_delay} observations.
\end{itemize}

Note that all methods above follow the procedure outlined in Section \ref{sec:workflow_experiments}. There are two differences between these approaches: (1) the degree of access to labels; and (2) how concept drift detection is carried out.

\subsection{Parameter Setup}\label{sec:parameter_setup}

In terms of parameters, we set the training window size $W$ to 1000 observations for most data streams. Due to low sample size, for the data streams \textit{Insects}, \textit{AbruptInsects}, \textit{Keystroke}, \textit{Ozone}, \textit{Outdoor}, and \textit{Luxembourg}, we set this parameter to 500 observations. We follow the setup used by Reis et al. \cite{dos2016fast} to set these values.

We focus on the Random Forest as learning algorithm \cite{breiman2001random}, which we apply with 100 trees. The remaining parameters are left as default according to the implementation provided in the \textit{scikit-learn} library \cite{pedregosa2011scikit}. We also apply the Random Forest method, with the same configuration, for building the student model in the proposed \texttt{STUDD} approach.

We apply the Page-Hinkley test \cite{page1954continuous} for concept drift detection, specifically its implementation from the \textit{scikit-multiflow} library \cite{montiel2018scikit}. This approach is a state of the art method for concept drift detection. We set the value of the $\delta$ parameter, which concerns the magnitude of changes, to 0.001, while the remaining parameters are left as default.

For the state of the art unsupervised concept drift detection approaches, the significance level parameter for rejecting the null hypothesis in the Kolmogorov-Smirnov test is set to 0.001, similarly to Reis et al. \cite{dos2016fast}. 
Regarding the delayed supervised methods (\texttt{DSS} and \texttt{DWS}), we set the delay parameter (\textit{l\_delay}) to $W / 2$, which is half of the training window size. For the weakly supervised variants (\texttt{WS} and \texttt{DWS}), the access to labels ((\textit{l\_access})) is set to 50. Finally, the loss function used as input to the Page-Hinkley test is the error rate.

\subsection{Results}\label{sec:experimental_results}

In this section, we present the results obtained from the experiments. First, we start by visualizing the alarms triggered by \texttt{STUDD} for concept drift and comparing to those of a supervised benchmark method (Section \ref{sec:visual_alarms}). Then, we present the main results which shows the performance of each approach and the respective costs (Section \ref{sec:main_results}). Finally, we carry out a sensitivity analysis which compares the performance of \texttt{STUDD} with a supervised approach with varying degrees of access to labels (Section \ref{sec:sensitivity_labels}).

\subsubsection{Visualizing Alarms}\label{sec:visual_alarms}

\begin{figure}[hbt]
\centering
\includegraphics[width=.9\textwidth]{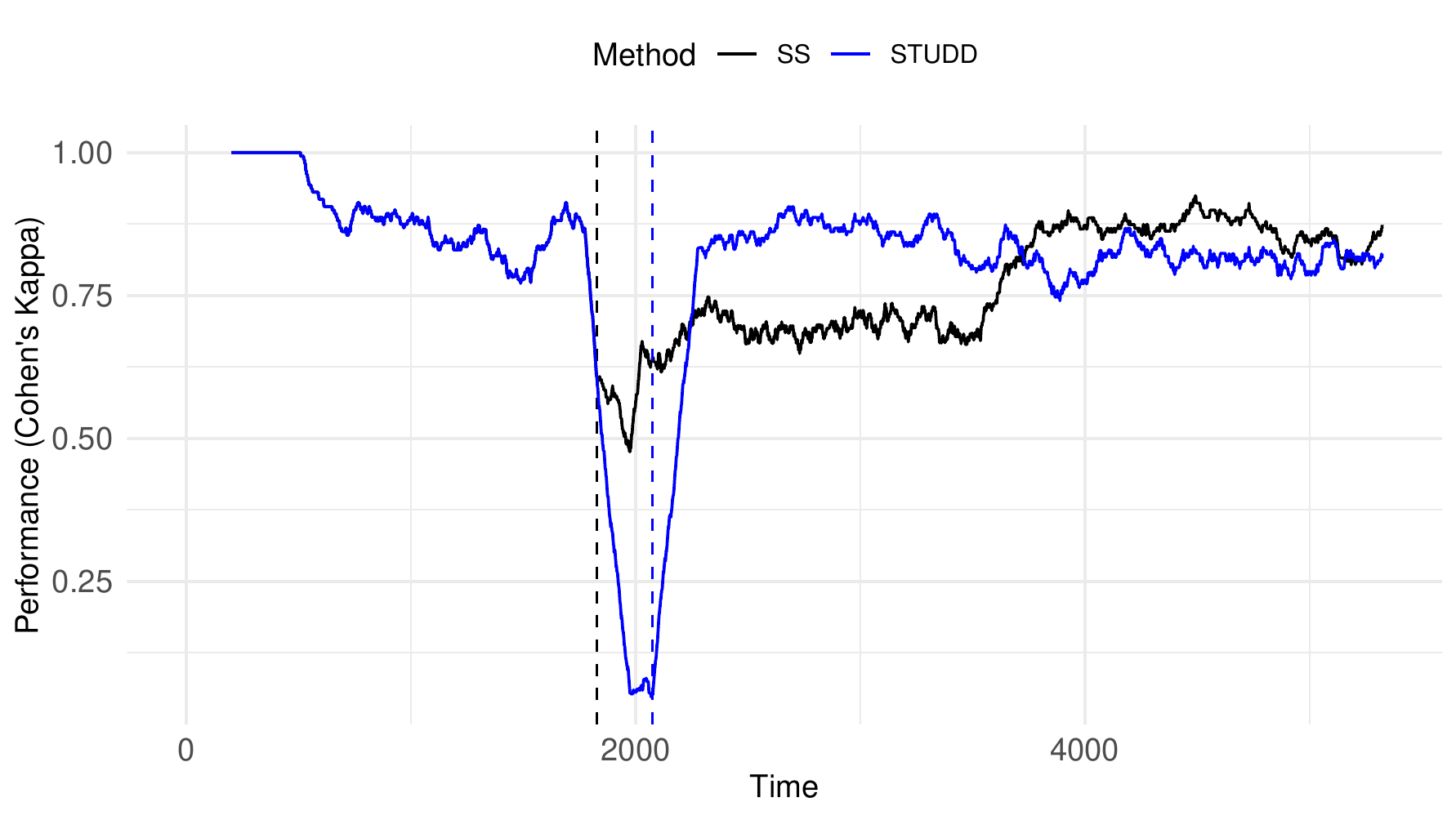}
\caption{An example using the \textit{AbruptInsects} data stream where the proposed method is able to detect concept drift and adapt to the environment similarly to a supervised approach.}
\label{fig:viz_alarm_succ}
\end{figure}

We start the analysis of the results by visualizing the alarms launched and the predictive performance by \texttt{STUDD}. We also include the behaviour of \texttt{SS} for a comparative analysis. In the interest of conciseness, we focus on three examples of of the 19 problems: a successful example, in which \texttt{STUDD} is able to detect concept drift and obtain a competitive performance with a supervised approach with complete access to the true labels; a positive example, in which the proposed approach shows a better change detection behaviour relative to \texttt{SS}; and a negative example, which shows a problem in which \texttt{STUDD} performs poorly.

\begin{figure}[hbt]
\centering
\includegraphics[width=.9\textwidth]{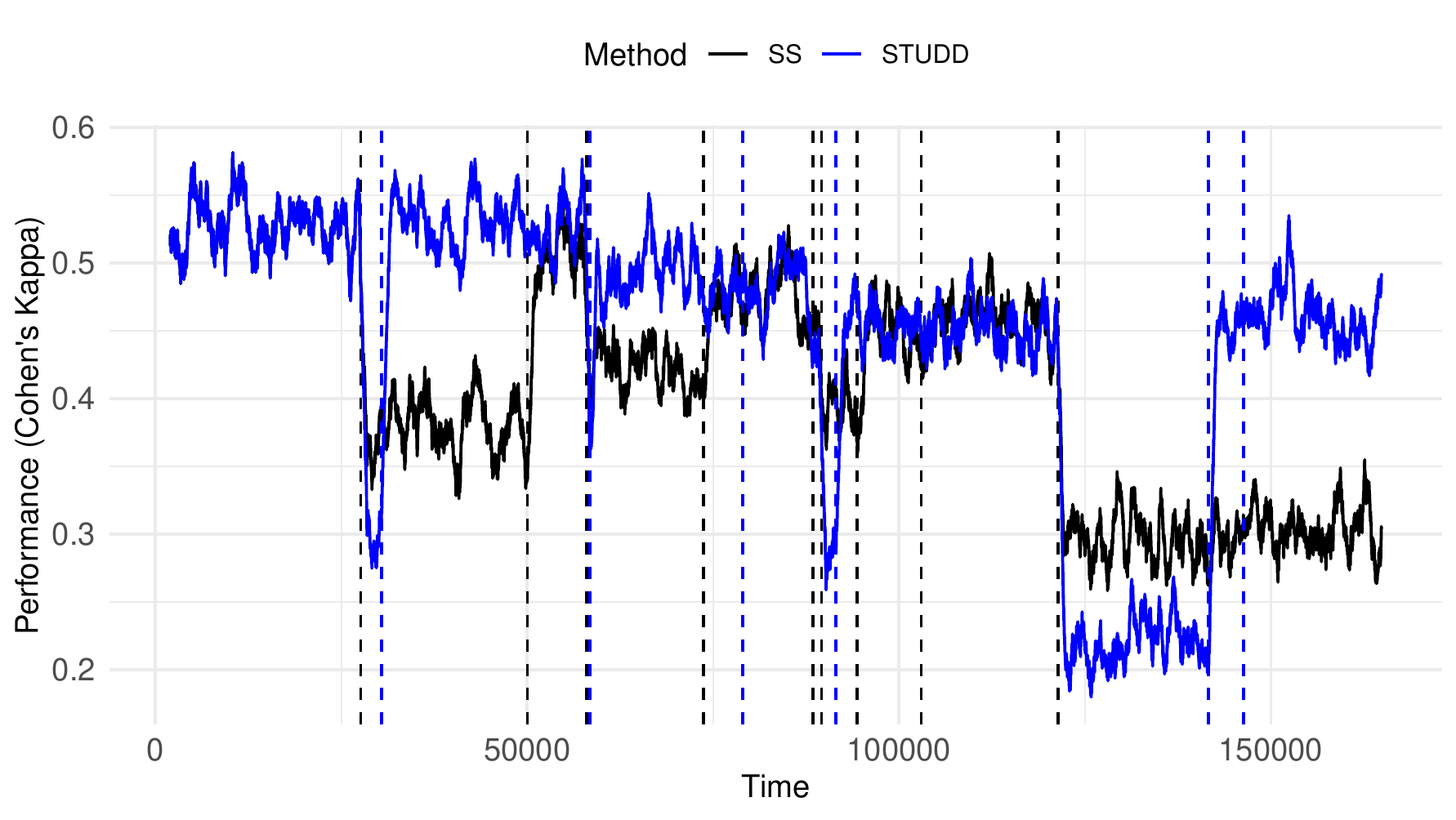}
\caption{An example using the \textit{Posture} data stream where the proposed method is able to detect concept drift and adapt to the environment better than a supervised approach.}
\label{fig:viz_alarm_positive}
\end{figure}

The first example is shown in Figure \ref{fig:viz_alarm_succ}. The figure shows the performance of each approach, \texttt{SS} in black and \texttt{STUDD} in blue, across the data stream \textit{InsectsAbrupt}. The performance is computed in a sliding window of 200 observations. The vertical dashed lines represent the time points in which the respective approach triggers an alarm for concept drift.

In the initial part of the data stream the performance of both approaches is identical (both lines are superimposed). Their behaviour are different from the point \texttt{SS} triggers the first alarm. This alarm has a visible impact on predictive performance because the score of \texttt{STUDD} decreases considerably. Notwithstanding, \texttt{STUDD} is able to detect the change soon after and regain the previous level of predictive performance.
This example shows that the proposed approach is able to detect changes in the environment.

\begin{figure}[hbt]
\centering
\includegraphics[width=.9\textwidth]{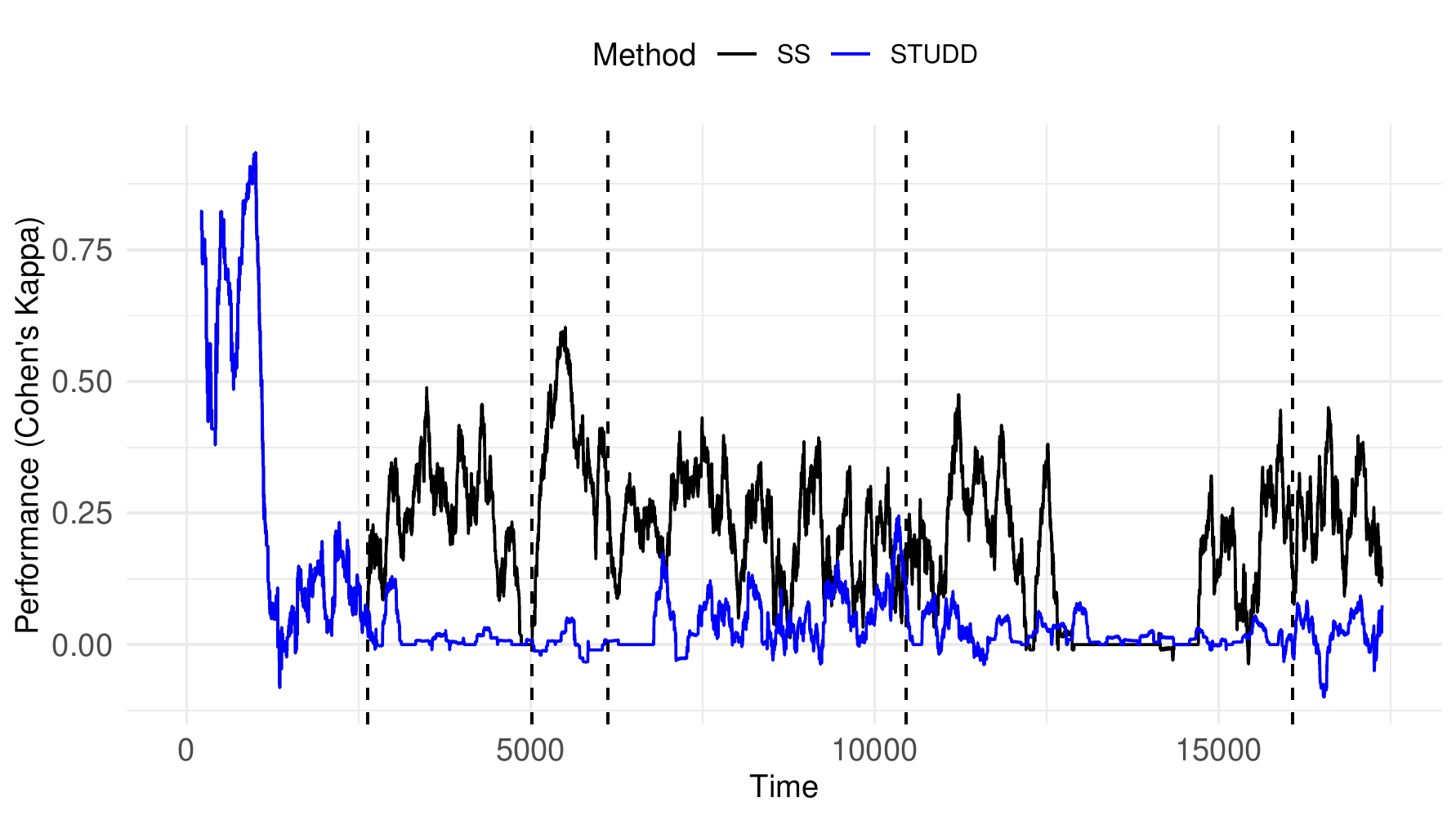}
\caption{An example using the \textit{Bike} data stream where the proposed method is unable to detect changes in the environment.}
\label{fig:viz_alarm_negative}
\end{figure}

Figure \ref{fig:viz_alarm_positive} shows another example for the data stream \textit{Posture} which follows the same structure as the previous one. In this case, \texttt{STUDD} is not only able to detect multiple changes in a timely manner but also show a visible better predictive performance relative to the benchmark. The proposed approach also launches less alarms relative to \texttt{SS}, which shows that it can be more efficient in terms of amount of labels it requests.

We show a final example in Figure \ref{fig:viz_alarm_negative} for data stream \textit{Bike}. This represents a negative example from the perspective of \texttt{STUDD}, in which it fails to detect the changes in the environment and performs poorly. On the contrary, the benchmark approach is able to improve its performance by detecting concept drift.

The above examples show the behaviour of \texttt{STUDD} is different scenarios.
In the next section, we will analyse its performance in all data streams and compare it to state of the art approaches.

\subsubsection{Performance by Data Stream}\label{sec:main_results}

The main results are presented in Tables \ref{tab:performance} and \ref{tab:costs}. The first one reports the Kappa score of each approach across each data set. The final row of the table described the average rank of each method across all problems. The second table has a similar structure as Table \ref{tab:performance}, but the values represent the ratio of labels (with respect to the full length of the data stream) used by the respective approach.

\texttt{STUDD} shows better performance scores relative to the static baseline \texttt{BL-st} (whose model is never updated) in most problems. Moreover, Table \ref{tab:costs} indicates that \texttt{STUDD} presents the best scores in terms of costs apart from the above-mentioned baseline. This outcome shows that the change signals triggered by \texttt{STUDD} are beneficial in terms of predictive performance, and that the method is efficient in terms of the labels required relative to other state of the art approaches.

\begin{table}
\label{tab:performance}
\caption{Performance of each method in each data set according to to Cohen's kappa score. The value of the best method in each category (baseline, supervised, and unsupervised) is in bold.}
\centering
\resizebox{\textwidth}{!}{%
\begin{tabular}[t]{lrrrrrrrrrr}
\toprule
  & \texttt{STUDD} & \texttt{BL-st} & \texttt{BL-ret} & \texttt{SS} & \texttt{DSS} & \texttt{WS} & \texttt{DWS} & \texttt{OS} & \texttt{OF} & \texttt{FF}\\
\midrule
AbruptInsects & 0.79 & 0.56 & 0.74 & 0.79 & 0.79 & 0.81 & 0.74 & 0.79 & 0.80 & 0.78\\
Insects & 0.81 & 0.81 & 0.77 & 0.77 & 0.76 & 0.81 & 0.81 & 0.70 & 0.71 & 0.76\\
Posture & 0.45 & 0.33 & 0.48 & 0.41 & 0.47 & 0.46 & 0.48 & 0.38 & 0.42 & 0.43\\
Arabic & 0.82 & 0.68 & 0.79 & 0.82 & 0.83 & 0.82 & 0.80 & 0.81 & 0.82 & 0.77\\
Bike & 0.02 & 0.02 & 0.33 & 0.31 & 0.31 & 0.35 & 0.28 & 0.02 & 0.02 & 0.33\\
\addlinespace
NOAA & 0.40 & 0.40 & 0.46 & 0.44 & 0.46 & 0.46 & 0.44 & 0.42 & 0.44 & 0.47\\
Sensor & 0.60 & 0.11 & 0.80 & 0.84 & 0.68 & 0.79 & 0.65 & 0.62 & 0.67 & 0.81\\
Powersupply & 0.06 & 0.06 & 0.08 & 0.05 & 0.07 & 0.03 & 0.03 & 0.07 & 0.08 & 0.08\\
Poker & 0.27 & 0.15 & 0.60 & 0.65 & 0.61 & 0.59 & 0.57 & 0.42 & 0.20 & 0.60\\
Rialto & 0.26 & 0.17 & 0.33 & 0.44 & 0.27 & 0.38 & 0.25 & 0.27 & 0.33 & 0.34\\
\addlinespace
Ozone & 0.13 & 0.13 & 0.12 & 0.13 & 0.13 & 0.13 & 0.13 & 0.13 & 0.13 & 0.13\\
Outdoor & 0.38 & 0.44 & 0.40 & 0.41 & 0.43 & 0.44 & 0.44 & 0.40 & 0.46 & 0.39\\
Luxembourg & 1.00 & 1.00 & 1.00 & 1.00 & 1.00 & 1.00 & 1.00 & 1.00 & 1.00 & 1.00\\
Gas & 0.32 & 0.38 & 0.54 & 0.63 & 0.58 & 0.69 & 0.54 & 0.53 & 0.41 & 0.54\\
Keystroke & 0.88 & 0.88 & 0.93 & 0.88 & 0.88 & 0.88 & 0.88 & 0.88 & 0.88 & 0.93\\
\addlinespace
ArabicShuffled & 0.83 & 0.83 & 0.84 & 0.83 & 0.83 & 0.83 & 0.83 & 0.84 & 0.83 & 0.83\\
Covtype & 0.43 & 0.39 & 0.70 & 0.70 & 0.59 & 0.65 & 0.51 & 0.26 & 0.53 & 0.71\\
GMSC & 0.10 & 0.10 & 0.20 & 0.24 & 0.14 & 0.10 & 0.10 & 0.10 & 0.10 & 0.10\\
Electricity & 0.40 & 0.28 & 0.52 & 0.49 & 0.49 & 0.43 & 0.48 & 0.51 & 0.46 & 0.52\\
\addlinespace
Avg. Rank & 6.97 & 7.82 & 4.16 & 4.34 & 4.50 & 4.32 & 6.03 & 6.58 & 5.79 & 4.50\\
\bottomrule
\end{tabular}%
}
\end{table}

We also compare the proposed approach to other unsupervised methods for concept drift detection, namely \texttt{OS}, \texttt{OF}, and \texttt{FF}. Regarding average rank, \texttt{STUDD} shows the best score cost-wise but the worst one in terms of performance. This result suggests that the proposed approach is more conservative relative to other unsupervised approaches, and better suited for dealing with false alarms. 
Overall, \texttt{FF} presents the best performance scores in terms of predictive performance, among the unsupervised approaches. However, these are accompanied with the worst scores in terms of costs, among all methods excluding the baseline \texttt{BL-ret}.
\texttt{OS} and \texttt{OF} present a more comparable behaviour relative to \texttt{STUDD}. While the performance average ran of \texttt{STUDD} is worse than that of \texttt{OS} and \texttt{OF}, the difference in scores are negligible in most of the problems.
On the other hand, the gains in predictive performance shown by \texttt{OS} and \texttt{OF} relative to \texttt{STUDD} come at a high cost. This is especially noteworthy in data sets \textit{Insects}, \textit{Poker}, \textit{Rialto}, \textit{Outdoor}, \textit{Gas}, and \textit{Electricity}, where \texttt{STUDD} shows relative low costs while maintaining comparable levels of performance. This suggests that the proposed approach is more efficient in terms of the number of labels required while securing a competitive performance.

\begin{table}
\label{tab:costs}
\caption{Ratio of additional labels (with respected to the full length of the data stream) required by each method in each data set. The value of the best method in each category (baseline, supervised, and unsupervised) is in bold.}
\centering
\resizebox{\textwidth}{!}{%
\begin{tabular}[t]{lrrrrrrrrrr}
\toprule
  & \texttt{STUDD} & \texttt{BL-st} & \texttt{BL-ret} & \texttt{SS} & \texttt{DSS} & \texttt{WS} & \texttt{DWS} & \texttt{OS} & \texttt{OF} & \texttt{FF}\\
\midrule
AbruptInsects & 0.19 & 0.09 & 0.94 & 0.19 & 0.19 & 0.19 & 0.19 & 0.19 & 0.19 & 0.66\\
Insects & 0.09 & 0.09 & 0.94 & 0.28 & 0.28 & 0.09 & 0.09 & 0.47 & 0.66 & 0.94\\
Posture & 0.04 & 0.01 & 0.99 & 0.06 & 0.06 & 0.16 & 0.08 & 0.02 & 0.02 & 0.74\\
Arabic & 0.23 & 0.11 & 0.91 & 0.34 & 0.23 & 0.23 & 0.68 & 0.23 & 0.23 & 0.80\\
Bike & 0.06 & 0.06 & 0.98 & 0.35 & 0.17 & 0.17 & 0.35 & 0.06 & 0.06 & 0.98\\
\addlinespace
NOAA & 0.11 & 0.06 & 0.99 & 0.22 & 0.28 & 0.22 & 0.28 & 0.17 & 0.17 & 0.99\\
Sensor & 0.57 & 0.01 & 0.99 & 1.22 & 0.48 & 1.71 & 0.87 & 0.39 & 0.45 & 0.99\\
Powersupply & 0.10 & 0.03 & 0.97 & 0.10 & 0.17 & 0.10 & 0.07 & 0.47 & 0.47 & 0.94\\
Poker & 0.02 & 0.01 & 0.99 & 0.78 & 0.55 & 1.03 & 0.75 & 0.32 & 0.21 & 0.99\\
Rialto & 0.10 & 0.01 & 1.00 & 1.49 & 0.58 & 1.32 & 0.68 & 0.50 & 0.78 & 1.00\\
\addlinespace
Ozone & 0.39 & 0.39 & 1.18 & 0.39 & 0.39 & 0.39 & 0.39 & 0.39 & 0.39 & 0.99\\
Outdoor & 0.38 & 0.25 & 1.00 & 0.75 & 0.75 & 0.25 & 0.25 & 0.62 & 0.75 & 1.00\\
Luxembourg & 0.53 & 0.53 & 1.05 & 0.53 & 0.53 & 0.53 & 0.53 & 0.53 & 0.53 & 1.05\\
Gas & 0.43 & 0.07 & 0.93 & 1.39 & 1.00 & 1.54 & 1.57 & 0.50 & 0.86 & 0.93\\
Keystroke & 0.31 & 0.31 & 0.94 & 0.31 & 0.31 & 0.31 & 0.31 & 0.31 & 0.31 & 0.63\\
\addlinespace
ArabicShuffled & 0.11 & 0.11 & 0.91 & 0.11 & 0.11 & 0.11 & 0.11 & 0.23 & 0.11 & 0.23\\
CovType & 0.12 & 0.01 & 0.99 & 0.65 & 0.35 & 0.69 & 0.19 & 0.11 & 0.31 & 0.99\\
GMSC & 0.01 & 0.01 & 0.99 & 0.01 & 0.01 & 0.01 & 0.01 & 0.01 & 0.01 & 0.01\\
Electricity & 0.18 & 0.02 & 0.99 & 0.73 & 0.44 & 0.51 & 0.35 & 0.44 & 0.53 & 0.99\\
\addlinespace
Avg. Rank & 3.50 & 2.13 & 9.24 & 6.24 & 5.21 & 5.84 & 5.16 & 4.34 & 4.76 & 8.58\\
\bottomrule
\end{tabular}%
}
\end{table}

When comparing \texttt{STUDD} with supervised approaches it is clear that, given our experimental setup, that having access to true labels brings an advantage for concept drift detection. All supervised variants, namely \texttt{SS}, \texttt{DSS}, \texttt{WS}, and \texttt{DWS}, show better performance relative to \texttt{STUDD}. Notwithstanding, the proposed approach is worthwhile in terms of costs, which are are considerably lower relative to these methods. For example, in the data streams \textit{Insects}, \textit{NOAA}, and \textit{Rialto}, \texttt{STUDD} is able to show a comparable performance with considerably less costs.



\subsubsection{Sensitivity Analysis to Label Access and Delay}\label{sec:sensitivity_labels}

In the previous two sections we showed the applicability of \texttt{STUDD} for concept drift detection, and how it compares with other state of the art approaches. While we approach the concept drift task in a completely unsupervised manner, there may be scenarios in which labels are available, though in a limited manner. We described these scenarios in Section \ref{sec:label_availability}. 
We introduced observation delay and availability in some of the methods used in the experiments, namely \texttt{DSS}, \texttt{WS}, and \texttt{DWS}. In this section, we aim at making another comparison between \texttt{STUDD} and these methods. Specifically, our goal is to study the relative performance of these methods for different values of label delay (\textit{l\_delay}) and label access (\textit{l\_access}).

As described in Section \ref{sec:methods}, we define the label availability according to two parameters: \textit{l\_access}\%, which denotes the probability of a label becoming available; and \textit{l\_delay}, which represents the number of observations it takes for a label to become available. 
For \textit{l\_access}, we test the following values: \{1, 10, 25, 50\}.
In terms of \textit{l\_delay}, the set of possibilities is \{250, 500, 1000, 1500, 2000, 4000\}.
For example, suppose that \textit{l\_access} is equal to 50 and \textit{l\_delay} is set to 250. This means that a label becomes available with 50\% probability after 250 observations.

\begin{figure}[hbt]
\centering
\includegraphics[width=\textwidth]{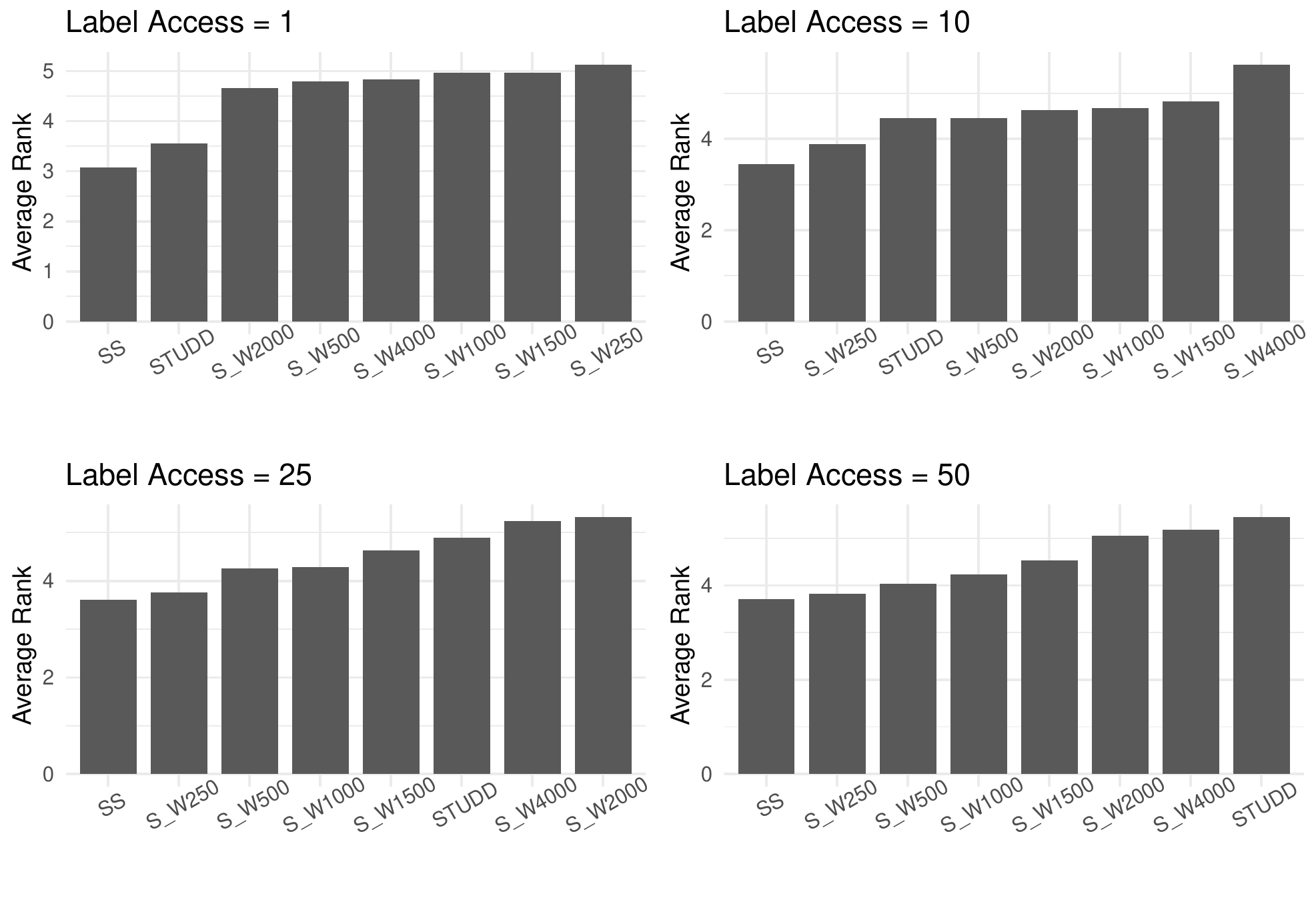}
\caption{Analysing the results for different values of \textit{l\_access} and \textit{l\_delay}. Each barplot represents the average rank of each approach across the 19 data streams.}
\label{fig:sens_analysis}
\end{figure}

We show the results of this analysis in Figure \ref{fig:sens_analysis}, which presents four barplots, one for each value of l\_access. Each barplot represents the average rank of each method across the 19 data streams. A method has rank 1 in a data set if it presents the best performance score in that task. In each barplot, we include \texttt{SS}, \texttt{STUDD}, and six supervised variants (one for each delay value). Each one of these methods is identified by the delay value. For example, \texttt{S\_W2000} represents a supervised variant with a delay of 2000 observations and the respective \textit{l\_access}. We note that in each barplot, the value of \textit{l\_access} is only valid for the six supervised variants and not for \texttt{SS} or \texttt{STUDD}.

The results show that \texttt{STUDD} performs relatively better as the probability of the label availability (\textit{l\_access}) decreases. Regarding the delay time (\textit{l\_delay}), lower values typically lead to better performance in terms of average rank. However, this parameter has a weaker impact relative to \textit{l\_access}. For example, for a \textit{l\_access} equal to 50, \texttt{STUDD} is the worst approach irrespective of the delay time. In summary, the results indicate that the proposed approach is beneficial if label acquisition is a problem.

\section{Discussion}\label{sec:discussion}

In the previous section we analysed the proposed approach for concept drift detection. \texttt{STUDD} is designed to detect changes in the environment without any access to true labels. In this sense, we refer to this approach as unsupervised.

The results obtained provided enough empirical evidence verifying the ability of \texttt{STUDD} to detect concept drift (\textbf{RQ1}). While its predictive performance is comparable to other unsupervised approaches in most of the problems, it is often able to considerably reduce the label requirements (\textbf{RQ2}). This feature is important in domains of application in which the annotation process or false alarms is costly. We also compared \texttt{STUDD} with several variants of supervised approaches to concept drift detection. The results indicated that \texttt{STUDD} provides better performance only if the access to labels is low (\textbf{RQ3}). 

In terms of future work, a possibly interesting research direction is attempting to combine \texttt{STUDD} with supervised approaches with potentially delayed or limited feedback.

\section{Conclusions}\label{sec:conclusions}

Detecting concept drift is an important task in predictive analytics. Most of the state of the art approaches designed to tackle this problem are based on monitoring the loss of the underlying predictive model. 

In this paper, we follow the idea that the assumption that labels are readily available for computing the loss of predictive models is too optimistic \cite{vzliobaite2010change,pinto2019automatic}. Therefore, we focus on solving this problem in an unsupervised manner, i.e., without any access to the true labels.

We propose a method to deal with this task based on a model compression \cite{bucilua2006model} approach. The core of the idea is to replace the loss of the predictive model which is deployed in the data stream (the teacher) with the \textit{mimicking} loss of the student model as the input to traditional concept drift detection methods, such as the Page-Hinkley test \cite{page1954continuous}. 

We carry out empirical experiments using 19 benchmark data streams and several state of the art methods. We show that the proposed method is able to detect concept drift and adapt itself to the environment. The behavior of the method is conservative with respect to other approaches, which is an advantage in domains where false alarms or label acquisition is costly. We published the code necessary to reproduce the experiments online. The data sets used are also available in public online repositories.

\bibliographystyle{spmpsci}      

\end{document}